\documentclass[runningheads]{llncs}
\usepackage{tabularx}
\usepackage{amsmath,amssymb}
\usepackage{colortbl}      % 行背景色依赖
\usepackage{xcolor}        % 颜色定义
\usepackage{booktabs}  
\usepackage[T1]{fontenc}

\usepackage{graphicx}
\usepackage[switch]{lineno}
\begin{document}
\title{TrafficRAG: A Multimodal RAG Framework for Traffic Accident Liability Determination}
\titlerunning{TrafficRAG}
%\author{First Author\inst{1}\orcidID{0000-1111-2222-3333} \and
%Second Author\inst{2,3}\orcidID{1111-2222-3333-4444} \and
% %Third Author\inst{3}\orcidID{2222--3333-4444-5555}}
% \author{Anonymous submission}
% %
% %\authorrunning{F. Author et al.}
% \authorrunning{Anonymous submission}
% %\institute{Princeton University, Princeton NJ 08544, USA \and
% %Springer Heidelberg, Tiergartenstr. 17, 69121 Heidelberg, Germany

% %\email{lncs@springer.com}\\
% %\url{http://www.springer.com/gp/computer-science/lncs} \and
% %ABC Institute, Rupert-Karls-University Heidelberg, Heidelberg, Germany\\
% %\email{\{abc,lncs\}@uni-heidelberg.de}}

\author{Xu Li \and ZeDong Fu \and Xinyi Li \and Xun Han}
\authorrunning{X. Li et al.}

% 单位、邮箱，区分两所院校
\author{
Xu Li\inst{1} \and
ZeDong Fu\inst{1} \and
Xinyi Li\inst{1} \and
Xun Han\inst{2} 
}
\authorrunning{X. Li et al.}

% \institute{
% \inst{1} Southwest Petroleum University, China\\
% \inst{2} Sichuan Police College, China\\
% \email{xul@swpu.edu.cn, donghedd206@gmail.com, xinyi0733@outlook.com, hldwxhx@163.com}

% }
\institute{
    Southwest Petroleum University, China\\
    \email{xul@swpu.edu.cn, donghedd206@gmail.com, xinyi0733@outlook.com}
    \and
    Sichuan Police College, China\\
    \email{hldwxhx@163.com}
}

\maketitle              % typeset the header of the contribution

\begin{abstract}
Traffic accident liability analysis is a critical but challenging task in intelligent transportation and legal assistance. Existing approaches are often inefficient, subjective, and inconsistent, while large language models are limited by noisy video inputs and insufficient legal grounding. To address these challenges, we propose \textit{TrafficRAG}, a multimodal retrieval-augmented framework to generate accident analysis reports. \textit{TrafficRAG} first uses a vision-language model to generate structured accident descriptions, which are used as retrieval queries. Given the text queries, a hybrid retrieval strategy combining BM25 sparse retrieval and dense retrieval is used to obtain relevant traffic regulations and similar case precedents. Finally, the large language model combines the retrieved knowledge and multimodal evidence to perform reasoning and generate a legally grounded liability analysis report. Extensive experiments demonstrate that our framework consistently outperforms several strong baselines, achieving 77.32\% in \textit{Legal Norm Adaptation Accuracy}, 81.71\% in \textit{Factual Faithfulness}, and 5.48\% in \textit{Liability Ratio MAE}. Our work demonstrates that a reasoning approach that combines multimodal case facts with legal clauses can significantly improve the performance of liability determination in traffic accidents. 

% First, the framework preprocesses raw accident videos into standardized representations. It then employs a vision-language model to generate structured accident descriptions, which are used as retrieval queries. Next, a hybrid retrieval strategy combining BM25 and dense retrieval is used to obtain relevant traffic regulations and similar case precedents. Finally, a generation model integrates multimodal evidence with retrieved knowledge to produce structured liability reports supported by explicit legal references. Experiments on a self-constructed multimodal traffic accident dataset show that TrafficRAG consistently outperforms several strong baselines, achieving 77.32\% in Legal Norm Adaptation Accuracy, 81.71\% in Factual Faithfulness, and 5.48\% in Liability Ratio MAE. These results demonstrate the effectiveness of integrating multimodal fact extraction with retrieval-augmented legal reasoning for traffic accident liability report generation.

\keywords{Retrieval-Augmented Generation \and Traffic Accident Liability Determination \and Multimodal Fusion.}
\end{abstract}
\section{Introduction}
	Traffic accident liability analysis is a critical task in intelligent transportation systems and judicial assistance \cite{lajmi2025towards,shen2025cadd}. Although the human-driven method of analyzing surveillance videos for liability determination remains the dominant approach, it is extremely inefficient, labor-intensive, and prone to inconsistent judgments when handling similar cases \cite{chen2024autonomous}. Establishing automated analysis methods enables quick determination of liability, which is crucial to improving road traffic flow and traffic management efficiency \cite{liu2022research}. Recently, multimodal AI has enabled vision-language models (VLMs) to achieve impressive video understanding and event description \cite{akter2025large,wu2024accidentgpt,zhang2025language}, and large language models (LLMs) to excel in legal reasoning and structured report generation \cite{huang2024chatgpt,ahmadi2025automatic,su2025judge}. VLM-based and LLM-based methods provide a promising training-free paradigm for improving the accuracy and efficiency of traffic accident liability determination. 
    % With the increasing number of road users and more complex road environments, traffic management authorities urgently require more efficient, consistent, and transparent accident processing \cite{liu2022research}.
    
    % Traditional manual procedures are still dominant but highly inefficient, labor-intensive, and prone to inconsistent decisions in similar cases \cite{liu2022research,chen2024autonomous}. Although recent progress in multimodal AI has enabled vision-language models to achieve impressive video understanding and event description \cite{akter2025large,wu2024accidentgpt,zhang2025language}, and large language models(LLMs) to excel in legal reasoning and structured report generation \cite{huang2024chatgpt,ahmadi2025automatic,su2025judge}, directly deploying such general models remains difficult.
    Existing VLM-based approaches typically follow a ``video perception $\rightarrow$ event description'' pipeline. Models including AccidentGPT \cite{wu2024accidentgpt}, TrafficVLM \cite{dinh2024trafficvlm}, and GPT-4V-based traffic assistants \cite{zhou2024gpt} excel at describing accident scenes. However, they focus on general visual understanding rather than liability-related fact extraction. In real-world surveillance scenarios, these models are vulnerable to noise, occlusion, and prompt variations \cite{akter2025large,fan2024learning}. In contrast, LLM-based methods adopt a ``text understanding $\rightarrow$ reasoning/report generation'' paradigm, performing well in logical reasoning and structured report generation \cite{huang2024chatgpt,ahmadi2025automatic,zhen2024leveraging,su2025judge}. Yet they rely heavily on upstream text quality and implicit knowledge, leading to unstable factual consistency and insufficient legal grounding for liability judgment \cite{abdelrahman2025advanced,ma2025sdd}.

    % Real-world traffic videos are often noisy and redundant, legal knowledge cannot be fully encoded in model parameters, and generated results must strictly guarantee factual accuracy and reliable legal grounding
    With the increasing complexity of the road environment, existing automated methods face several challenges: 1) traffic surveillance videos typically exhibit substantial noise and redundancy; 2) legal knowledge cannot be exhaustively represented within model parameters; and the generated analysis results struggle to ensure factual correctness and robust, verifiable legal justification \cite{shen2025cadd,su2024stard,ma2025sdd}. To bridge these challenges, we propose \textit{TrafficRAG}, a multimodal retrieval-augmented  framework that processes accident videos, extracts structured descriptions via VLMs, and integrates external legal knowledge and precedents to generate standardized, legally grounded liability determination reports \cite{lewis2020retrieval,yang2025transrag,zou2026marag}.
	
	The main contributions of this paper are as follows:
    
        $\bullet$ We offer an overall multimodal dataset for traffic accident liability analysis.
        
		$\bullet$ We propose \textit{TrafficRAG}, a multimodal retrieval-augmented framework for traffic accident liability determination.
		% \item We introduce a cross-source consistency reranking module that jointly selects mutually supportive legal provisions and analogous cases, improving the coherence of retrieved evidence for downstream report generation.
		% \item We design a liability-oriented accident description paradigm and a dual-path retrieval mechanism that combines BM25 and dense retrieval to support legal grounding and case-based reasoning.
        
		  $\bullet$ We validate the effectiveness of \textit{TrafficRAG} through extensive experiments and analyze the contribution of different components to performance through ablation studies.
	
	The remainder of this paper is organized as follows. Section 2 reviews the related work. Section 3 introduces the proposed \textit{TrafficRAG} framework. Section 4 presents the experimental setup, dataset construction, evaluation results, and ablation studies, followed by a discussion of limitations and ethical considerations. Section 5 concludes the paper and describes future research directions.
	
	\section{Related Work}
	\subsection{Vision-Language Models and Multimodal Traffic Accident Analysis}
    Deep learning-based multimodal accident analysis forms the basis of traffic accident liability determination. Early CNN-based methods mainly focused on scene recognition or accident detection, but showed limited ability to capture the temporal evolution of accident events \cite{fang2023vision,melegrito2024deep}. With the development of Transformer-based architectures, temporal modeling has become more effective for extracting accident behavior sequences and dynamic event information \cite{singh2025surveillance,dinh2024trafficvlm}.
    
    More recently, VLMs have opened new directions for multimodal traffic accident understanding. TrafficVLM improves traffic scene understanding and event description through enhanced controllability \cite{dinh2024trafficvlm}, while GPT-4V shows strong generalization in complex traffic event understanding \cite{zhou2024gpt}. Several studies have also reformulated traffic accident analysis as language-centric tasks for identifying collision causes and risk factors \cite{akter2025large,fan2024learning,zhen2024leveraging,abdelrahman2025advanced}. However, most existing methods focus on general scene understanding or event description, rather than extracting liability-relevant facts for legal reasoning.
    
    \subsection{Legal Retrieval Augmentation and Judicial Document Generation}
    The accurate use of legal knowledge is essential for liability report generation. In legal retrieval and text generation, methods such as SAILER and Lawformer improve case retrieval and long-text semantic modeling \cite{li2023sailer,xiao2021lawformer}, respectively, but they are primarily designed for pure-text input and are not directly applicable to multimodal accident scenarios derived from traffic videos.
    
    Existing legal text generation approaches generally rely on LLMs for legal prediction and reasoning \cite{su2025judge,wu2023precedent}, while methods such as SDD-LawLLM further enhance performance through task-specific fine-tuning \cite{ma2025sdd}. However, these methods depend heavily on implicit parametric knowledge and lack explicit external legal grounding. Retrieval-augmented generation (RAG) addresses this limitation by combining external retrieval with text generation \cite{lewis2020retrieval}. Recent methods such as TransRAG and MaRAG have extended RAG to traffic, maritime, and aviation accident analysis \cite{yang2025transrag,zou2026marag,ren2025retrieval}. However, these approaches are not tailored for liability-oriented accident descriptions extracted from multimodal videos.
Thus, they cannot fully support the closed-loop process of ``fact extraction $\rightarrow$ knowledge retrieval $\rightarrow$ liability report generation''.
This research gap motivates our proposed \textit{TrafficRAG} framework.
	
	\section{Method}
	\subsection{Overall Framework}
	This work addresses the multi-stage task of generating accident liability determination reports for accident videos. Given a video $V$, the preprocessing module $P$ obtains a unified representation, and the VLM $M_v$ produces a structured accident description $x$. Using $x$ as the query, the legal provision retriever $R_{\text{law}}$ and case retriever $R_{\text{case}}$ obtain candidates $\mathcal{L}^{\text{cand}}$ and $\mathcal{C}^{\text{cand}}$. A cross-source consistency reranking module $R_{\text{joint}}$ selects a compact evidence bundle $\mathcal{B}$, and the generation model $G$ outputs the final report $y$.

    The overall pipeline is formalized as follows:
    \begin{equation}
    \setlength{\arraycolsep}{2pt}
    \begin{cases}
        x = M_v(P(V)), \quad
        \mathcal{L}^{\text{cand}} = R_{\text{law}}(x), \quad
        \mathcal{C}^{\text{cand}} = R_{\text{case}}(x), \\[3pt]
        \mathcal{B} = R_{\text{joint}}(x,\mathcal{L}^{\text{cand}},\mathcal{C}^{\text{cand}}), \quad
        y = G(x,\mathcal{B}).
    \end{cases}
    \end{equation}
    
    The generated report $y$ includes basic accident information, progression, legal basis, liability apportionment, reasoning, and conclusion. By unifying retrieval and reranking of legal provisions and cases, our framework improves evidence coherence for report generation.
	
	\subsection{Video Preprocessing and VLM-Based Accident Description Generation}
\begin{figure}[htbp]
        \centering
        \includegraphics[width=\textwidth]{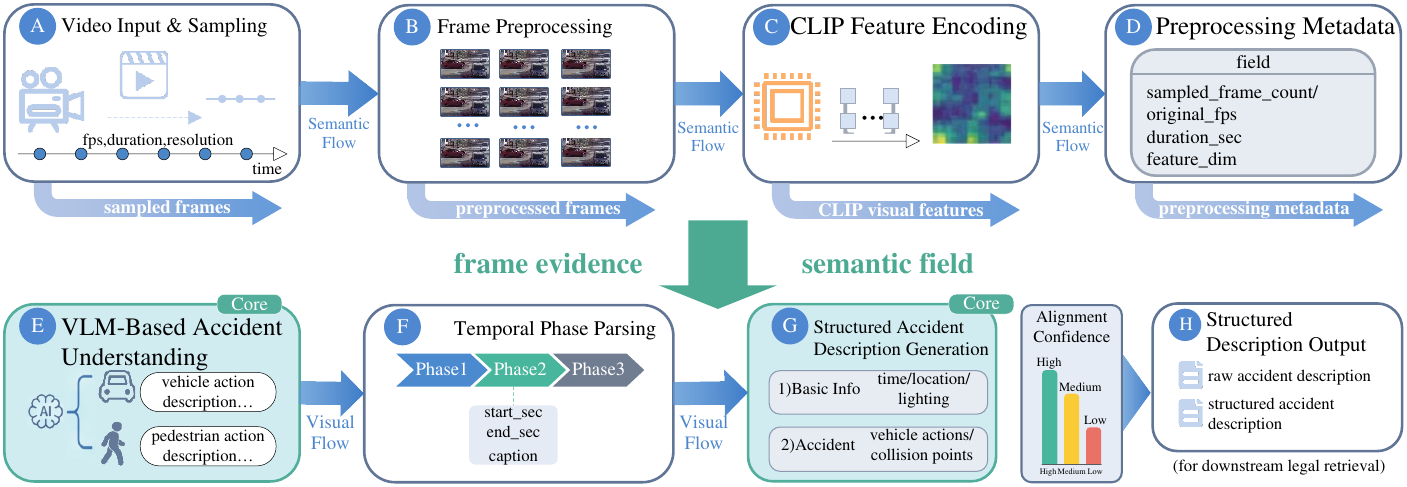}
        \caption{Video preprocessing and structured accident description extraction pipeline.}
        \label{fig1}
    \end{figure}

    The video preprocessing module $P$ (which handles frame sampling, denoising, and feature encoding)aims to reduce noise in raw traffic accident videos, standardize the input to the VLM, and preserve the temporal continuity of accident events. As shown in Fig.~\ref{fig1}, $P$ first samples frames from the raw video $V$ at a fixed frame rate to remove redundancy. The sampled frames are then denoised and normalized, after which the frame sequence is encoded into dense visual features using CLIP (\texttt{openai/clip-vit-large-patch14}). To balance temporal coverage and input compactness, the system retains a limited set of key frames and maps them into a unified visual representation space.
    
    To generate accident descriptions for liability determination, we design $M_v$ with a visual-encoder--language-decoder architecture. As illustrated in Fig.~\ref{fig1}, the model first produces a raw accident description and then refines it into a structured, liability-oriented description. The visual encoder extracts scene and motion features, including interactions among traffic participants, while the language decoder maps these features into a natural-language description $x$.
    
    To focus the model on liability-relevant information, we formulate this stage as a multimodal generation task guided by prompts. Specifically, $M_v$ is guided to extract four types of information: scene background, participants, behavioral sequence, and liability-related cues. The resulting structured description $x$ preserves key accident facts and serves as the query for subsequent legal retrieval and evidence selection.
    
    \subsection{Dual-Path Knowledge Retrieval with Cross-Source Consistency Reranking}
    Given the liability-oriented accident description $x$, we retrieve relevant legal provisions and similar cases through a dual-path hybrid retriever that combines BM25 sparse retrieval with dense vector retrieval. This design balances lexical matching and semantic similarity. For dense retrieval, we use \texttt{moka-ai/m3e-base} as the encoder and build the index with FAISS.
    
    For a query $x$ and a candidate document $d$, we compute the sparse score $S_{\text{sparse}}(x,d)$ and the dense score $S_{\text{dense}}(x,d)$. After Min--Max normalization over the top-$K$ candidates, the retrieval score is defined as
    \begin{equation}
        S_{\text{ret}}(x,d)=\alpha \tilde{S}_{\text{dense}}(x,d) + (1-\alpha)\tilde{S}_{\text{sparse}}(x,d),
    \end{equation}
    where $\alpha \in [0,1]$ controls the trade-off between dense and sparse retrieval.
    
    To improve evidence coherence, we further introduce a cross-source consistency reranking module. For each candidate law-case pair $(l_i,c_j)$, the joint score is
    \begin{equation}
        S_{\text{joint}}(l_i,c_j \mid x)=
        \lambda_1 S_{\text{fact}}(x,c_j)+
        \lambda_2 S_{\text{norm}}(x,l_i)+
        \lambda_3 S_{\text{cons}}(l_i,c_j),
    \end{equation}
    where $S_{\text{fact}}(x,c_j)$ is the cosine similarity between m3e-base embeddings of the accident description and the retrieved case, $S_{\text{norm}}(x,l_i)$ uses the same encoder to assess applicability of the legal provision to the accident description, and $S_{\text{cons}}(l_i,c_j)$ is scored via fixed logic rules based on liability consistency (with no learned parameters or cross-encoder). The weights $\lambda_1,\lambda_2,\lambda_3$ and threshold $\tau$, detailed in Section 4.1, are tuned on the validation set. The top-ranked pairs are retained as the final evidence bundle $\mathcal{B}$.
    
    \subsection{Liability Determination Report Generation Module}
\begin{figure}[htbp]
      \centering
      \includegraphics[width=\textwidth]{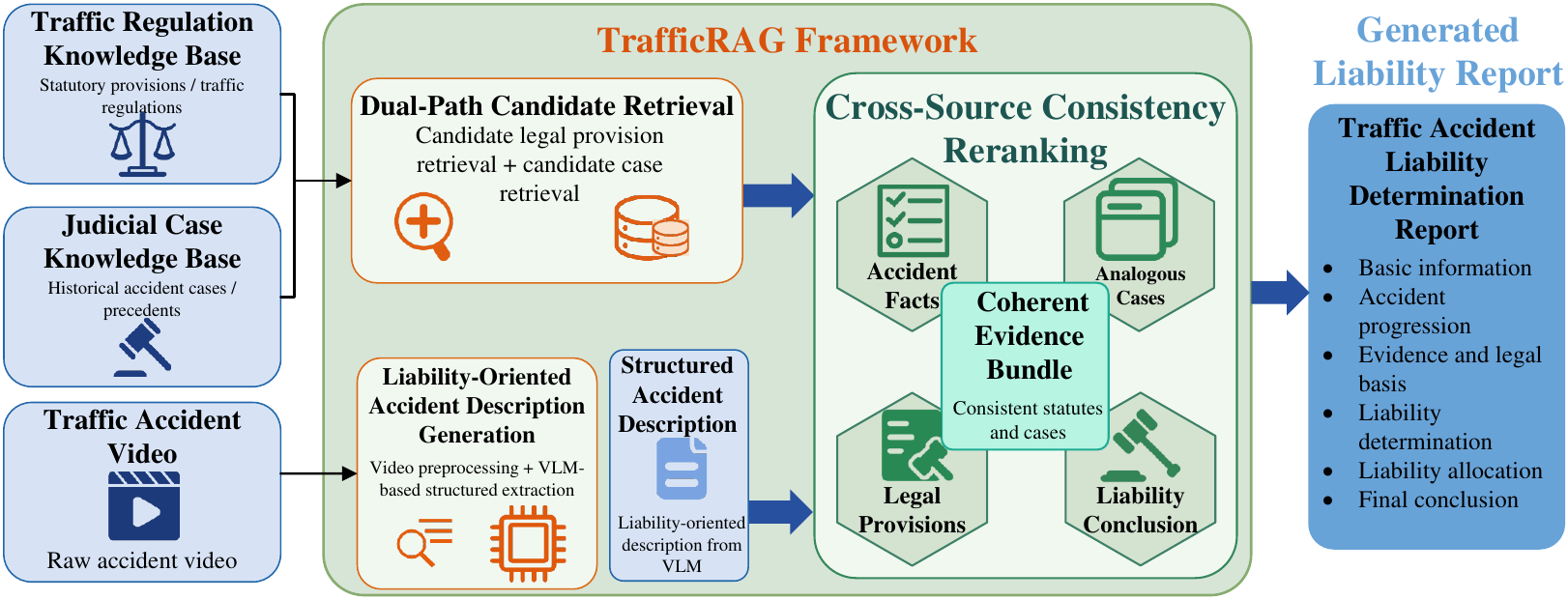}
      \caption{Liability determination report generation pipeline.}
      \label{fig2}
    \end{figure}

    We formulate liability report generation as a conditional text generation task under external knowledge constraints. As shown in Fig.~\ref{fig2}, the generator uses a structured prompt to produce reports with standardized sections, including basic accident information, accident progression, evidence and legal basis, liability determination, liability allocation, and the final conclusion.
    
    The generation objective is
    \begin{equation}
      \max_{\theta} \sum_{t=1}^{T} \log p_{\theta}(y_t \mid y_{<t}, x, \mathcal{B}),
    \end{equation}
    where $y_t$ is the $t$-th token in the target report, $\theta$ denotes the model parameters, $x$ is the structured accident description, and $\mathcal{B}$ is the reranked evidence bundle.
    
    To reduce hallucinations and improve controllability, we adopt structured prompting and require the generated report to ground its legal reasoning and liability conclusions in $\mathcal{B}$. During generation, we enforce three forms of consistency: factual consistency between the report and the accident facts, legal consistency between the conclusions and the retrieved evidence, and structural consistency across different sections of the report.
	
	\section{Experiments}
	\subsection{Experimental Setup and Training}
	\subsubsection{Dataset Construction}
    We employ a standardized curation pipeline to construct a dedicated multimodal traffic accident dataset, integrating public accident understanding benchmarks with legal resources for liability attribution and case retrieval. Video data are collected from VRU-Accident\cite{kim2025vru} and TAU-106K\cite{zhou2025tau}, while legal resources are assembled from CADD\cite{shen2025cadd}, STARD\cite{su2024stard}, and LeCaRDv2\cite{li2024lecardv2}. Candidate videos are filtered by visual quality and annotation validity before being combined with metadata and preliminary VLM-generated descriptions into a unified annotation framework.

    Annotators refine the automated descriptions into liability-focused structured representations and complete standardized reference reports that include applicable legal provisions, liability allocation, and evidence keywords. Ambiguous cases with unclear facts or insufficient legal support are carefully verified, revised, or excluded to ensure data reliability.
    
    The final dataset comprises 1,584 traffic accident cases, with each sample containing raw accident videos, metadata, structured descriptions, official liability reports, and supporting legal information. We split the dataset at the case level to form disjoint training, validation, and test sets, preventing sample overlap and retrieval leakage.
    
    \subsubsection{Knowledge Base Construction}
    To support retrieval-augmented generation, we build two external knowledge bases: a legal knowledge base and a case knowledge base. The legal knowledge base includes 265 national traffic regulation provisions related to liability determination, collected from the Road Traffic Safety Law and its Implementation Regulations. The knowledge base consists of 671 traffic accident cases, each with accident descriptions, liability judgments, reasoning, penalties, and legal references. Public judicial documents were standardized, filtered, and deduplicated, and overlapping cases were excluded to avoid retrieval leakage.
    
    \subsubsection{Training and Hyperparameter Settings}
    In video preprocessing, raw videos are uniformly sampled at 3 FPS and encoded by CLIP ViT-L/14. For the vision-language component, we adopt an InternVL2-style 8B backbone as $M_v$ and fine-tune it with LoRA for liability-oriented accident description generation, using rank 64, learning rate $2\times10^{-4}$, batch size 16, and 3 epochs. The output is constrained to four fields: scene background, participants, behavioral sequence, and liability-related cues.
    
    For retrieval, we combine BM25 with \texttt{m3e-base} dense retrieval and use FAISS for vector indexing, with sparse and dense scores fused by Eq.~(2) using $\alpha=0.6$. In the cross-source consistency reranking stage, the weights $\lambda_1,\lambda_2,\lambda_3$ and threshold $\tau$ are selected on the validation set. For report generation, we use \texttt{Qwen3-Max} with a unified template to produce liability determination reports.
	
	\subsection{Evaluation Metrics}
    To evaluate the full pipeline from video input to liability report generation, we assess performance from four aspects: \emph{evidence coverage}, \emph{legal retrieval quality}, \emph{legal-factual consistency}, and \emph{liability ratio prediction accuracy}.
    
    We use five metrics: Key Evidence Recall (KER) for evidence coverage \cite{su2025judge}, Statute Recall (SR) for legal provision retrieval or citation \cite{su2024stard}, Factual Faithfulness (FF) for factual support from accident facts and retrieved knowledge \cite{kryscinski2020evaluating,es2024ragas}, Legal Norm Adaptation Accuracy (LNA) for whether the cited provisions support the final liability conclusion \cite{yu2025benchmarking}, and Liability Ratio Mean Absolute Error (LR-MAE) for liability ratio prediction. Higher values indicate better performance for KER, SR, FF, and LNA, while lower values indicate better performance for LR-MAE. Detailed symbol definitions and formulas are provided in Appendix~\ref{app:metrics_detail}.
	
	\subsection{Overall Result Analysis}
	We compare \textit{TrafficRAG} with several baselines for legal reasoning and retrieval-augmented generation. Since most baselines are designed for text input rather than raw videos, all methods are evaluated under a unified video-to-text setting. To reduce confounding factors, we use the same test set, retrieval pool size, and report structure constraints for all methods whenever applicable.
	
	\begin{table}
		\centering
		\caption{Performance comparison of different methods on the traffic accident liability determination task.}
		\label{tab:performance_comparison}
		\begin{tabular}{lccccc}
			\hline
			Method & KER (\%) & SR (\%) & LNA (\%) & FF (\%) & LR-MAE (\%) \\
			\hline
			BM25(2024)      & --    & 34.00 & 22.81 & --    & --    \\
			QLD(2024)       & --    & 34.41 & 23.60 & --    & --    \\
			SAILER(2023)    & --    & 72.33 & 46.69 & --    & --    \\
			LawRAG(2024)    & 74.17 & 70.03 & 55.83 & 53.65 & 34.22 \\
            Judge(2025)     & 66.67 & 42.00 & 67.91 & 69.97 & 25.89 \\
			\hline
			DeepSeek-V3.2   & 77.31 & 81.83 & 72.41 & 72.41 & 10.74 \\
			Gemini-3.1-Pro  & 72.30 & 69.81 & 64.24 & 76.99 & 7.31  \\
			\hline
			\rowcolor[RGB]{206,234,247}
			TrafficRAG      & 82.87 & 84.79 & 77.32 & 81.71 & 5.48 \\
			\hline
		\end{tabular}
	\end{table}

    Table~\ref{tab:performance_comparison} shows that \textit{TrafficRAG} achieves the strongest overall results among the compared methods, obtaining the best scores on KER (82.87\%), SR (84.79\%), LNA (77.32\%), and FF (81.71\%), together with the lowest LR-MAE (5.48\%). These results suggest that combining liability-oriented fact extraction with consistency aware legal evidence selection improves fact coverage, legal grounding, and liability allocation in traffic accident liability report generation.
	
	\subsection{Ablation Study Analysis}
    To investigate the contributions of external knowledge and consistency-aware evidence selection, we conduct ablation experiments on the two retrieval sources, i.e., the legal knowledge base (Legal KB) and the case knowledge base (Case KB), as well as the proposed cross-source consistency re-ranking module.

    \begin{table}
    	\centering
    	\caption{Ablation study results on the retrieval sources and the cross-source consistency reranking module.}
    	\label{tab:ablation}
    	\begin{tabular}{ccc|ccccc}
    		\hline
    		\textbf{Legal KB} & \textbf{Case KB} & \textbf{Rerank} & KER (\%) & SR (\%) & LNA (\%) & FF (\%) & LR-MAE (\%) \\
    		\hline
    		$\times$ & $\times$ & $\times$ & 51.87 & 45.42 & 43.19 & 59.44 & 31.81 \\
    		$\times$ & $\checkmark$ & $\times$ & 75.28 & 62.73 & 45.91 & 73.65 & 15.69 \\
    		$\checkmark$ & $\times$ & $\times$ & 67.54 & 74.88 & 62.37 & 79.26 & 23.58 \\
    		$\checkmark$ & $\checkmark$ & $\times$ & 81.46 & 82.11 & 69.68 & 79.02 & 10.36 \\
    		\hline
    		\rowcolor[RGB]{206,234,247}
    		$\checkmark$ & $\checkmark$ & $\checkmark$ & 82.87 & 84.79 & 77.32 & 81.71 & 5.48 \\
    		\hline
    	\end{tabular}
    \end{table}
    
    As shown in Table~\ref{tab:ablation}, removing both retrieval sources yields the weakest overall performance, indicating that accident descriptions alone are insufficient for reliable liability determination. Using only the Case KB substantially reduces LR-MAE and improves FF, suggesting that similar cases support practical reasoning and liability allocation, although SR and LNA remain limited without explicit legal grounding. In contrast, using only the Legal KB markedly improves SR and LNA, confirming the importance of statutory support, but liability ratio prediction remains weaker than that of the full model.

    Using both knowledge sources without reranking provides complementary legal and analogical evidence, though the retrieved statutes and cases may lack strict alignment. In contrast, by enabling cross-source consistency reranking, the model achieves its best performance, demonstrating that mutually supportive law-case pairs effectively refine both liability conclusions and allocation.
    
    These results highlight two forms of complementarity: the Legal KB provides explicit normative grounding, the Case KB contributes practical analogical evidence, and the reranking module further strengthens their interaction by filtering weakly aligned law-case pairs.
    
    \begin{figure}[htbp]
    	\centering
    	\includegraphics[width=\textwidth]{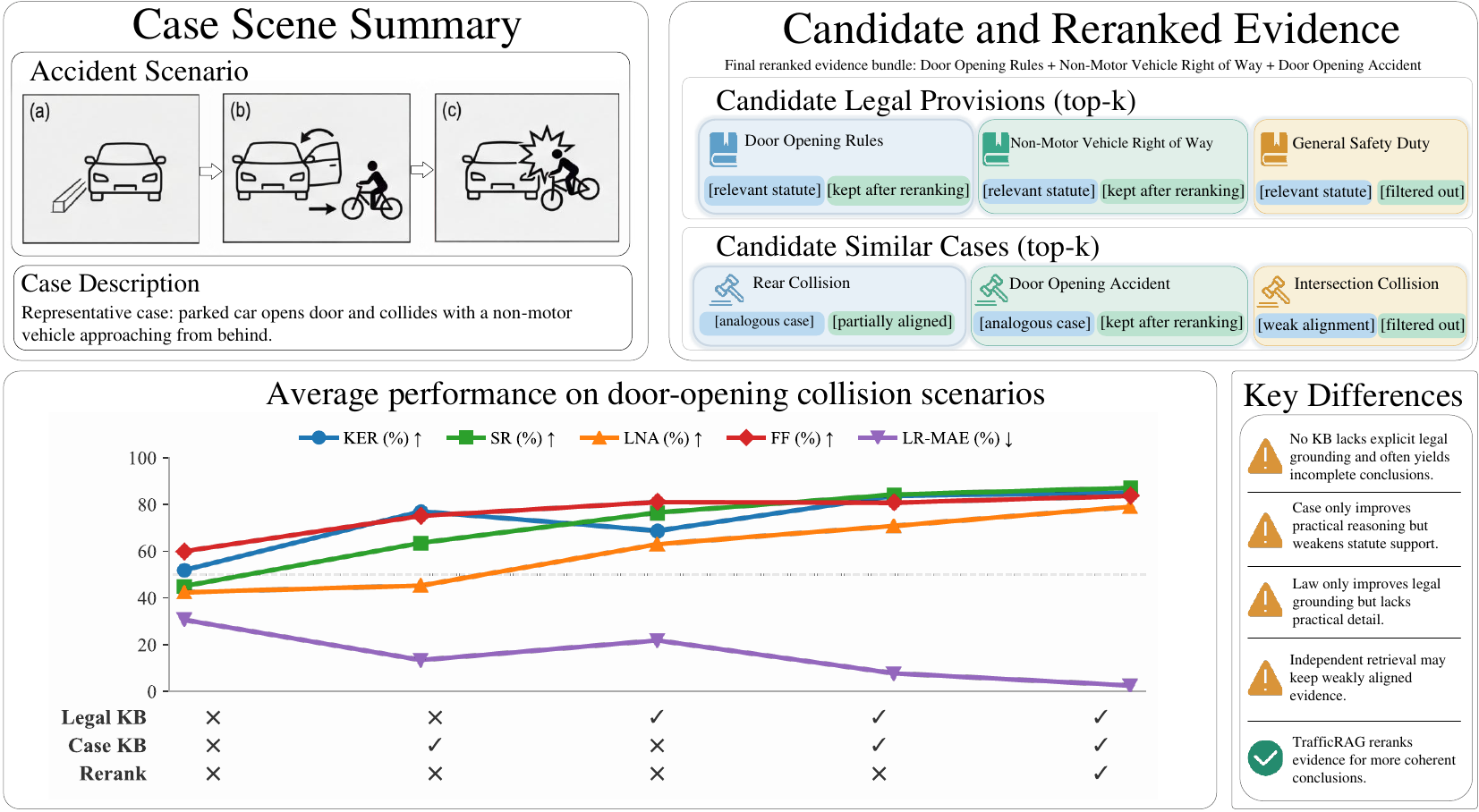}
    	\caption{Qualitative analysis of a representative door-opening collision scenario. TrafficRAG selects more coherent legal and case evidence and produces liability conclusions closer to the reference determination.}
        \label{fig4}
    \end{figure}
    
    To further illustrate this effect, we analyze a representative door-opening collision in which an opened car door strikes a following non-motor vehicle. As shown in Fig.~\ref{fig4}, direct generation without external knowledge tends to produce vague conclusions and incomplete legal support. Using only the Legal KB improves legal grounding but lacks practical detail, whereas using only the Case KB yields more plausible handling logic but weaker statutory support. By combining both sources with cross-source consistency reranking, \textit{TrafficRAG} produces reports that better match the reference determination in terms of liability conclusion, legal basis, and liability allocation.

   \section{Limitations and Ethical Considerations}
   TrafficRAG is a legal assistance prototype rather than a substitute for traffic administrators or legal experts, and its outputs require human review to be legally valid. Currently limited to Chinese traffic laws and cases, it has limitations such as video understanding error propagation, retrieval bias, unstable legal applicability, and potential biases from historical cases or incomplete evidence. We addressed these issues by anonymizing sensitive information and annotating legal articles with jurisdiction and time validity. Future work includes cross-jurisdictional validation, human-in-the-loop supervision, and impact evaluation. In short, TrafficRAG is an auxiliary tool instead of an autonomous decision system.
	
	\section{Conclusion and Future Work}
    In this paper, we proposed \textit{TrafficRAG}, a multi-stage framework for accident liability determination report generation. Experimental results show that the proposed framework consistently outperforms multiple baseline methods on the benchmark task. In future work, we plan to enhance the multimodal understanding and evidence reasoning capabilities, introduce stricter rule-based legal verification and confidence calibration mechanisms, and optimize the cross-source consistency modeling to further improve the reliability and interpretability of the framework for real-world traffic accident liability determination.

	\appendix
	\section{Notation Symbols and Evaluation Metrics}\label{app:metrics_detail}
    Let the $i$-th sample be indexed by $i$. We denote by $E_i$ and $\hat{E}_i$ the gold and extracted liability-related evidence sets, by $S_i$ and $\hat{S}_i$ the gold and generated cited statute sets, by $r_i$ and $\hat{r}_i$ the gold and generated liability conclusions, and by $a_i^{(j)}$ and $\hat{a}_i^{(j)}$ the gold and predicted liability proportions of the $j$-th liable party, respectively. Moreover, $\hat{F}_i$ denotes the set of extracted factual claims, $\mathbb{I}(c)$ is an indicator function, $m_i$ is the number of liable parties in the $i$-th sample, and $N$ is the total number of samples.
    
    Table~\ref{tab:metrics_formulas} summarizes the evaluation metrics used in this work.
    
    \begin{table*}
        \centering
        \caption{Detailed Formulas of Evaluation Metrics}
        \label{tab:metrics_formulas}
        \begin{tabular}{l l}
            \toprule
            \textbf{Metric} & \textbf{Formula} \\
            \midrule
            Key Evidence Recall (KER) &
            \begin{tabular}[c]{@{}l@{}}
            $\mathrm{KER} = \frac{1}{N}\sum_{i=1}^{N}\frac{|E_i \cap \hat{E}_i|}{|E_i|}$
            \end{tabular} \\
            \midrule
            Statute Recall (SR) &
            \begin{tabular}[c]{@{}l@{}}
            $\mathrm{SR} = \frac{1}{N}\sum_{i=1}^{N}\frac{|S_i \cap \hat{S}_i|}{|S_i|}$
            \end{tabular} \\
            \midrule
            Legal Norm Adaptation Accuracy \\(LNA) &
            \begin{tabular}[c]{@{}l@{}}
            Sample-level:
            $\mathrm{LNA}_i$\\
            $=\begin{cases}
            1, & \text{if } (\hat{S}_i \cap S_i \neq \varnothing)\wedge(\hat{r}_i = r_i), \\
            0, & \text{otherwise}
            \end{cases}$ \\
            Overall: $\mathrm{LNA} = \frac{1}{N}\sum_{i=1}^{N}\mathrm{LNA}_i$
            \end{tabular} \\
            \midrule
            Factual Faithfulness (FF) &
            \begin{tabular}[c]{@{}l@{}}
            $\mathrm{FF} = \frac{1}{N}\sum_{i=1}^{N}\frac{1}{|\hat{F}_i|}\sum_{c\in\hat{F}_i}\mathbb{I}(c)$
            \end{tabular} \\
            \midrule
            Liability Ratio Mean Absolute Error \\(LR-MAE) &
            \begin{tabular}[c]{@{}l@{}}
            $\mathrm{LR\text{-}MAE} = \frac{1}{N}\sum_{i=1}^{N}\frac{1}{m_i}\sum_{j=1}^{m_i}\left|\hat{a}_i^{(j)} - a_i^{(j)}\right|$
            \end{tabular} \\
            \bottomrule
        \end{tabular}
    \end{table*}

	\bibliographystyle{splncs04}
	\bibliography{mybib}

\end{document}